\documentclass{article}

\PassOptionsToPackage{square,numbers}{natbib}

\usepackage[preprint]{neurips_2023}

\usepackage[utf8]{inputenc} 
\usepackage[T1]{fontenc}    
\usepackage{booktabs}       
\usepackage{tabularx} 
\usepackage{amsfonts}       
\usepackage{nicefrac}       
\usepackage{microtype}      
\usepackage{xcolor}         
\usepackage{amsmath}
\usepackage{amssymb} 
\usepackage{bm} 
\usepackage{amsthm} 
\usepackage{xfrac}
\usepackage{multirow}  
\usepackage{enumerate} 
\usepackage{enumitem} 
\usepackage{pgfplots} 
\usepackage{caption} 
\usepackage{subcaption} 
\usepackage[breaklinks=true,bookmarks=false]{hyperref}
\usepackage{url}
\usepackage{siunitx} 
\usepackage{soul}

\hypersetup{
    colorlinks = true,
    urlcolor = blue,
    citecolor=.,
    linkcolor=.,
}
\usepackage{cleveref}
\usepackage{tikz} 
\usetikzlibrary{positioning}
\usetikzlibrary{fadings}

\newcolumntype{Y}{>{\centering\arraybackslash}X}

\usepackage[colorinlistoftodos]{todonotes}

\def\rva{{\mathbf{a}}}

\def\rvc{{\mathbf{c}}}

\def\rvm{{\mathbf{m}}}

\def\rvv{{\mathbf{v}}}

\def\rvx{{\mathbf{x}}}

\def\rvz{{\mathbf{z}}}

\DeclareMathAlphabet{\mathsfit}{\encodingdefault}{\sfdefault}{m}{sl}
\SetMathAlphabet{\mathsfit}{bold}{\encodingdefault}{\sfdefault}{bx}{n}

\title{GAIA-1: \linebreak A Generative World Model for  Autonomous Driving}

\author{%
  Anthony Hu\textsuperscript{*}\\
  \And
  Lloyd Russell\textsuperscript{*}\\
  \And
  Hudson Yeo\textsuperscript{*}\\
  \And
  Zak Murez\\
  \And
  George Fedoseev\\
  \And
  Alex Kendall\\
  \And
  Jamie Shotton\\
  \And
  Gianluca Corrado\\
  \AND
  Wayve\\
  \texttt{research@wayve.ai}\\
  \textsuperscript{*} equal contributions
}

\begin{document}
\graphicspath{{Figures/}}

\maketitle

\begin{abstract}
    Autonomous driving promises transformative improvements to transportation, but building systems capable of safely navigating the unstructured complexity of real-world scenarios remains challenging. A critical problem lies in effectively predicting the various potential outcomes that may emerge in response to the vehicle's actions as the world evolves.

    To address this challenge, we introduce GAIA-1 (`Generative AI for Autonomy'), a generative world model that leverages video, text, and action inputs to generate realistic driving scenarios while offering fine-grained control over ego-vehicle behavior and scene features. Our approach casts world modeling as an unsupervised sequence modeling problem by mapping the inputs to discrete tokens, and predicting the next token in the sequence. Emerging properties from our model include learning high-level structures and scene dynamics, contextual awareness, generalization, and understanding of geometry. The power of GAIA-1’s learned representation that captures expectations of future events, combined with its ability to generate realistic samples, provides new possibilities for innovation in the field of autonomy, enabling enhanced and accelerated training of autonomous driving technology.
\end{abstract}

\section{Introduction}
\label{section:introduction}

\begin{figure}[t]
    \centering
    \includegraphics[width=\linewidth]{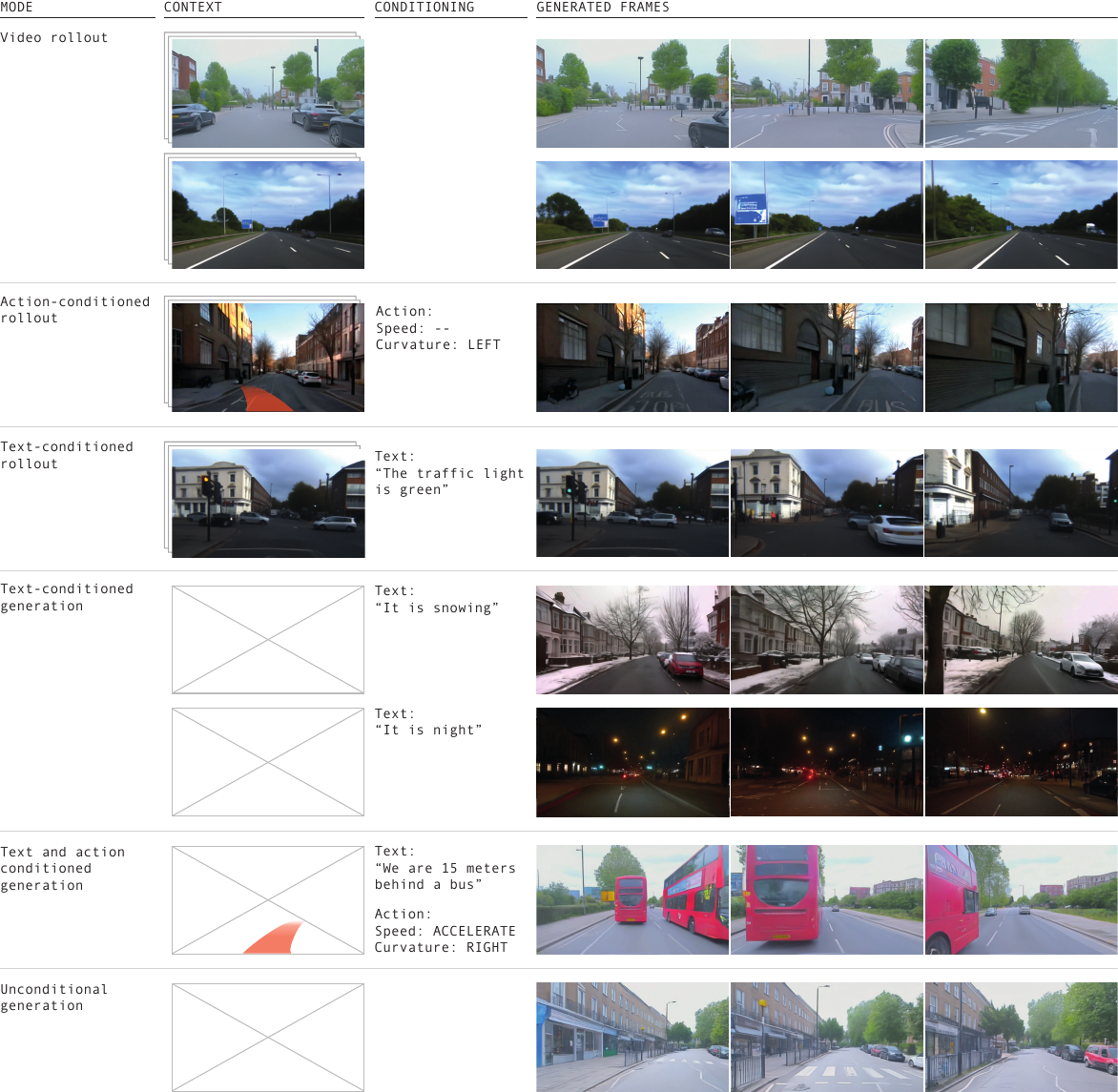}%
    \caption{GAIA-1 multimodal video generation. GAIA-1 can generate videos by performing future rollouts starting from a video prompt. These future rollouts can be further conditioned on actions to influence particular behaviors of the ego-vehicle (e.g. steer left), or on text to drive a change in some aspects of the scene (change the color of the traffic light). For speed and curvature we condition the model by passing the sequence of future speed and / or curvature values. Our model can also generate realistic videos from text prompts, or by simply drawing samples from its prior distribution (fully unconditional generation).} 
    \label{fig:prediction_task}
\end{figure}

Predicting future events is a fundamental and critical aspect of autonomous systems. Accurate future prediction enables autonomous vehicles to anticipate and plan their actions, enhancing safety and efficiency on the road. To achieve this, the development of a robust model of the world is imperative \citep{kendall2019learning} and huge efforts have been made in the past to build such predictive world models for autonomous driving~\citep{nuscenes19,hu2020probabilistic, waymo21,hu2021fiery,hu2023_uniad}. A world model \citep{ha18,lecun22} learns a structured representation and understanding of the environment that can be leveraged for making informed decisions when driving.

However, current approaches have had significant limitations. World models have been successfully applied to control tasks in both simulation~\cite{schrittwieser2020mastering,janner2021sequence,hu2022mile,micheli2023transformers,hafner2023mastering} and to real-world robotics tasks~\cite{reed22gato,wu2023daydreamer}. These methods often rely on labeled data, which is challenging to obtain at scale, and models that work on simulated data may not fully capture the complexities of real-world scenarios. Furthermore, due to their low-dimensional representations, these models may struggle to generate highly realistic samples of future events, posing challenges in achieving a high level of fidelity in predictions for complex real-world applications such as autonomous driving.

Meanwhile, progress in generative image and video generation has harnessed the power of self-supervised learning to learn from large quantities of real-world data to generate remarkably realistic video samples~\citep{ho2022imagen,harvey2022flexible,esser2023structure}. Yet, a significant challenge persists in this domain: the difficulty of learning a representation that captures the expected future events. While such generative models excel at generating visually convincing content, they may fall short in learning representations of the evolving world dynamics that are crucial for precise future predictions and robust decision-making in complex scenarios. 

In this work we introduce GAIA-1, a method designed with the goal of maintaining the benefits of both world models and generative video generation. It combines the scalability and realism of generative video models with the ability of world models to learn meaningful representations of the evolution into the future. GAIA-1 works as follows. First, we partition the model into two components: the world model and the video diffusion decoder. The world model reasons about the scene's high-level components and dynamics, while the diffusion model takes on the responsibility of translating latent representations back into high-quality videos with realistic detail.

For the world model, we use vector-quantized representations of video frames to discretize each frame, transforming them into a sequence of tokens. Subsequently, we reframe the challenge of predicting the future into predicting the next token in the sequence \citep{janner2021sequence,chen2021decisiontransformer}. This approach has been widely employed in recent years to train large language models \citep{megatron22,Chowdhery2022PaLMSL,llama23,openai2023gpt4}, and it is recognized for its effectiveness in enhancing model performance through the scaling of model size and data. It is possible to generate samples within the latent space of the world model through autoregressive generation.

The second component is a multi-task video diffusion decoder that is able to perform high-resolution video rendering as well as temporal upsampling to generate smooth videos from the information autoregressively generated by the world model. Similarly to large language models, video diffusion models have demonstrated a clear correlation between scale of training and overall performance, making both components of GAIA-1 suitable for effective compound scaling.

GAIA-1 is designed to be multimodal, allowing video, text and action to be used as prompts to generate diverse and realistic driving scenarios, as demonstrated in Figure~\ref{fig:prediction_task}. By training it on a large corpus of real-world UK urban driving data, GAIA-1 learns to understand and disentangle important concepts such as static and dynamic elements, including cars, buses, pedestrians, cyclists, road layouts, buildings, and even traffic lights. Further, it provides fine-grained control over both ego-vehicle behavior and other scene features through action and language conditioning.

GAIA-1 demonstrates the ability to manifest the generative rules of the real world. Emerging properties such as learning high-level structures, generalization, creativity, and contextual awareness indicate that the model can comprehend and reproduce the rules and behaviors of the world. Moreover, GAIA-1 exhibits understanding of 3D geometry, for example, by effectively capturing the intricate interplay of pitch and roll induced by road irregularities such as speed bumps. It showcases reactive behaviors of other agents demonstrating the ability to understand causality in decision making of road users. Surprisingly, it shows the capability to successfully extrapolate beyond the training data, for example to driving outside of the boundaries of the road. See \Cref{section:emerging-properties} for a comprehensive list of examples.

The power of GAIA-1's learned representations to predict future events, paired with control over both ego-vehicle dynamics and scene elements, is an exciting advance that paves the way for improving embodied intelligence and providing synthetic data to accelerate training and validation. World models, such as GAIA-1, are the basis for the ability to predict what might happen next, which is fundamentally important for decision-making in autonomous driving.

\section{Model}
\label{section:model-architecture}

\begin{figure}[t]
    \centering
    \includegraphics[width=\linewidth]{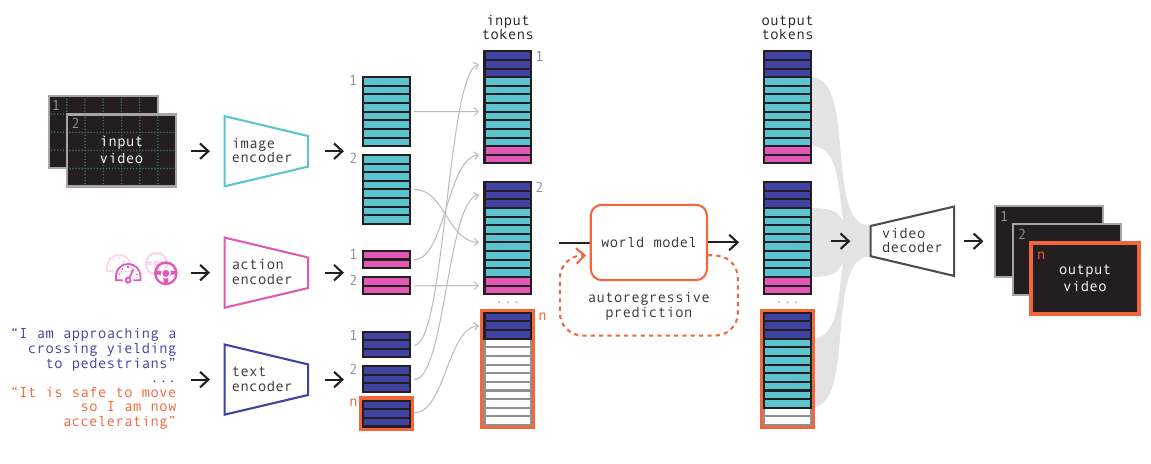}%
    \caption{Architecture of GAIA-1. First, we encode information from all input modalities (video, text, action) into a common representation: images, text and actions are encoded as a sequence of tokens. The world model is an autoregressive transformer that predicts the next image token conditioned on past image, text, and action tokens. Finally, the video decoder maps the predicted image tokens back to the pixel space, at a higher temporal resolution.} 
    \label{fig:architecture}
\end{figure}

In this section we describe the model architecture of the trainable components of GAIA-1. The general architecture is presented in \Cref{fig:architecture}.

\subsection{Encoding Video, Text and Action}
\label{subsection:encoding}
GAIA-1 can leverage three different input modalities (video, text, action), which are encoded into a shared $d$-dimensional space.

\paragraph{Image tokens.}Each image frame of a video is represented as discrete tokens. To achieve this, we use a pre-trained image tokenizer for discretization (for details about the pre-training see Section~\ref{subsection:image-tokenizer}). Formally, let us consider a sequence of $T$ images $(\rvx_1, \ldots, \rvx_T)$, where each image $\rvx_t$ in this sequence is discretized into $n=576$ discrete tokens using the pre-trained image tokenizer. We obtain a sequence denoted as $(\rvz_1, \ldots, \rvz_T)$, where each $\rvz_t = (z_{t,1}, \ldots, z_{t,n}) \in \mathbb{R}^n$ corresponds to $n = \frac{H}{D} \times \frac{W}{D}$ discrete tokens. Here, $H$ and $W$ represent the height and width of the input image, while $D$ denotes the downsampling factor of the image tokenizer. These discrete tokens are then mapped to a $d$-dimensional space via an embedding layer that is trained alongside the world model. 

\paragraph{Text tokens.} At each time step $t$, we incorporate information from both text and action. Textual input is encoded using the pre-trained T5-large model \citep{2020t5}, resulting in $m=32$ text tokens per time step. These tokens are mapped to a $d$-dimensional space through a linear layer that is trained in conjunction with the world model. This process yields a text representation denoted as $\rvc_t = (\rvc_{t,1}, \ldots, \rvc_{t,m}) \in \mathbb{R}^{m\times d}$.

\paragraph{Action tokens.} For actions, we consider $l=2$ scalar values (representing speed and curvature). Each scalar is independently mapped to the $d$-dimensional space via a linear layer that is trained with the world model. Consequently, the action at time step $t$ is represented as $\rva_t = (\rva_{t,1}, \ldots, \rva_{t,l}) \in \mathbb{R}^{l\times d}$. 

For each time step, the input tokens are interleaved in the following order: text - image - action. The final input of the world model is therefore $(\rvc_1,\rvz_1, \rva_1, \ldots, \rvc_T, \rvz_T, \rva_T)$. To encode the position of the input tokens, we use a factorized spatio-temporal positional embedding. 1) A learnable temporal embedding is shared across all the tokens of a given time step, i.e. there are $T$ temporal embeddings. 2) A learnable spatial embedding indicates the position of a token within a time step, i.e. there are $m + n + l = 610$ spatial embeddings ($m$ text tokens, $n$ image tokens, and $l$ action tokens) of dimension $d=4096$.

\subsection{Image Tokenizer}
\label{subsection:image-tokenizer}
When modeling discrete input data with a sequence model, there is a trade-off between the sequence length and the vocabulary size. The sequence length refers to the number of discrete tokens that are needed to describe the data. The vocabulary size corresponds to the number of possible values a single token can take. For language, there are two obvious choices for tokens: characters and words. When using character-level tokens, the input data has a longer sequence length, and each individual token belongs to a smaller vocabulary, but conveys little meaning. When using word-level tokens, the input data has a shorter sequence length, and each token contains a lot of semantics but the vocabulary is extremely large. Most language models \citep{roberta19,brown20,2020t5,Chowdhery2022PaLMSL,hoffmann22,llama23} use byte-pair encoding (or equivalent) as a trade-off between character-level and word-level tokenization.

Likewise for video, we would like to reduce the sequence length of the input, while possibly making the vocabulary larger, but with tokens that are more semantically meaningful than raw pixels. We do this with a discrete image autoencoder \citep{oord17}. There are two objectives we would like to achieve in this first stage:

\begin{enumerate}
    \item Compress the information from raw pixels to make the sequence modeling problem tractable. Images contain a lot of redundant and noisy information. We would like to reduce the sequence length needed to describe the input data.
    \item Guide the compression towards meaningful representations, such as semantics, instead of high-frequency signals. The resulting input space for the world model will be simpler to compose with, and less dominated by high-frequency signals that can considerably slow down the learning process.
\end{enumerate}

\begin{figure}[t]
    \centering 
\begin{subfigure}{0.33\textwidth}
  \includegraphics[width=\linewidth]{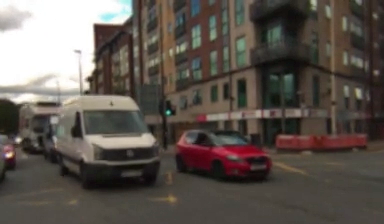}
  \caption{Input image}
  \label{fig:tokenizer_a}
\end{subfigure}\hfil 
\begin{subfigure}{0.33\textwidth}
  \includegraphics[width=\linewidth]{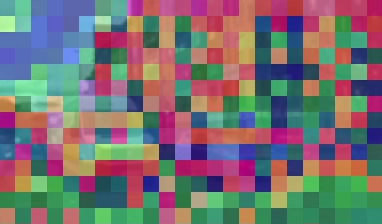}
  \caption{Base VQ-GAN tokens}
  \label{fig:tokenizer_b}
\end{subfigure}\hfil
\begin{subfigure}{0.33\textwidth}
  \includegraphics[width=\linewidth]{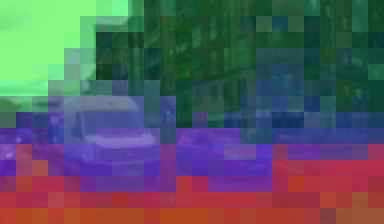}
  \caption{DINO-distilled tokens}
  \label{fig:tokenizer_c}
\end{subfigure}
\caption{Increasing semantic content of image tokens through DINO distillation. Visualization shows the top 3 PCA components of token embeddings mapped to RGB values. DINO-distilled tokens corresponding to a semantic class (e.g. vehicle, road, or sky) have similar embeddings.}
\label{fig:dino_tokenizer}
\end{figure}

We reduce the sequence length of the input data by downsampling each input image by a factor $D=16$ in both height and width. Each image $\rvx_t$ of size $H\times W$ is described by $n=\frac{H}{D} \times \frac{W}{D}$ tokens with a vocabulary size $K$. Inspired by~\citep{peng2022beit}, we guide the compression towards meaningful representations by regressing to the latent features of a pre-trained DINO model \citep{caron2021emerging}, a self-supervised image model that is known to contain semantic information. See \Cref{fig:dino_tokenizer} for a qualitative example.

The discrete autoencoder is a fully convolutional 2D U-Net \citep{ronneberger15}. The encoder $E_{\theta}$ quantizes the image features using nearest neighbor look-up from a learnable embedding table \cite{oord17}, resulting in image tokens $\rvz_t = E_{\theta}(\rvx_t)$.  Note that the decoder is only used to train the image autoencoder, solely the discrete encoder $E_{\theta}$ is part of the final GAIA-1 model. Due to the decoder being trained on single images it lacks temporal consistency when decoding to a video. For this reason we also train a video decoder that is described in \Cref{subsection:video-decoder}. 

The training losses for the image autoencoder are the following:
\begin{itemize}
\item \textbf{Image reconstruction loss.} The image reconstruction loss is a weighted sum of $L_1$, $L_2$, perceptual loss $L_{\text{perceptual}}$ \citep{Johnson2016Perceptual}, and GAN loss $L_{\text{GAN}}$ \citep{esser21}.

\item \textbf{Quantization loss.} To update the embedding vectors, we use the embedding loss and the commitment loss from \citep{oord17}. We adopted the linear projection of the embedding and $L_2$ normalization from \citep{yu2022vectorquantized} as we found this helped increase vocabulary usage.

\item \textbf{Inductive bias loss.} The quantized image features are encouraged to match the image features of a pre-trained DINO \citep{caron2021emerging} model with a cosine similarity loss. Distilling the information from DINO into the learned tokens is important as it allows them to benefit from the inductive biases of this model.
\end{itemize}

\subsection{World Model}
\label{subsection:world-model}
As described in \Cref{subsection:encoding} the input of the world model is $(\rvc_1,\rvz_1, \rva_1, ..., \rvc_T, \rvz_T, \rva_T)$. The world model is an autoregressive transformer network that models the sequence input. Its training objective is to predict the next image token in the sequence conditioned on all past tokens, using causal masking in the attention matrix of the transformer blocks \citep{vaswani17}.

\begin{equation}
L_{\text{world model}} = -\sum_{t=1}^T \sum_{i=1}^n\log p(z_{t,i} | \rvz_{<t}, z_{t, j < i}, \rvc_{\leq t}, \rva_{<t})
\end{equation}

We randomly dropout conditioning tokens during training so that the world model can do (i) unconditional generation, (ii) action-conditioned generation, and (iii) text-conditioned generation.

To further reduce the sequence length of our world model we temporally subsample videos from 25Hz to 6.25Hz. This allows the world model to reason over longer periods without leading to intractable sequence lengths. To recover video predictions at full frame rate we perform temporal super-resolution using the video decoder described in \Cref{subsection:video-decoder}.


\subsection{Video Decoder}
\label{subsection:video-decoder}
Following the recent advances in image \citep{rombach2022highresolution, saharia2022photorealistic} and video generation \citep{ho2022imagen, esser2023structure} we use denoising video diffusion models for the GAIA-1 decoder. A naive approach of independently decoding each frame-tokens to pixel space results in a temporally inconsistent video output. Modeling the problem as denoising a sequence of frames during the diffusion process, where the model can access information across time, greatly improves temporal consistency of the output video. 

We follow \citep{ho2022video} and use a 3D U-Net with factorized spatial and temporal attention layers. During training, our video diffusion model is conditioned on the image tokens obtained by discretizing input images with the pre-trained image tokenizer $E_{\theta}$. During inference, the diffusion model is conditioned on the predicted image tokens from the world model.

We train a single model jointly on both image and video generation tasks. Training on videos teaches the decoder to be temporally consistent, while training on images is crucial for the quality of individual frames \citep{ho2022imagen} as it teaches the model to extract information from conditioning image tokens. We disable temporal layers when training on images.

\begin{figure}
\centering
\begin{tabular}{cc}
\subfloat[Image generation]{\includegraphics[width = 1.75in]{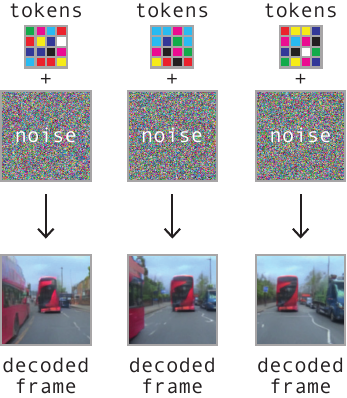}} \hspace{30pt} &
\subfloat[Video generation]{\includegraphics[width = 1.75in]{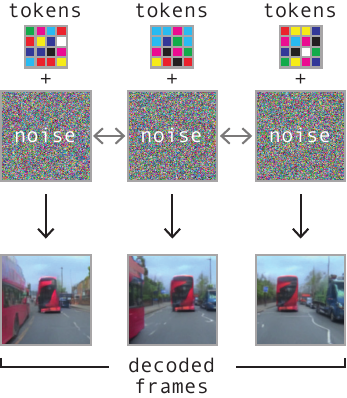}} \\ [30pt] 
\subfloat[Autoregressive video generation]{\includegraphics[width = 1.75in]{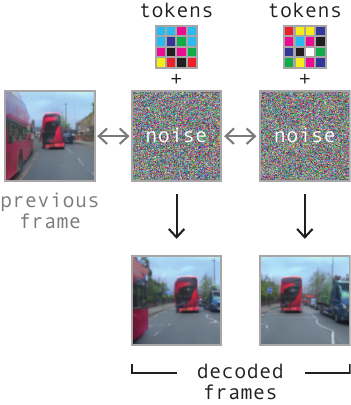}} \hspace{30pt} &
\subfloat[Video interpolation]{\includegraphics[width = 1.75in]{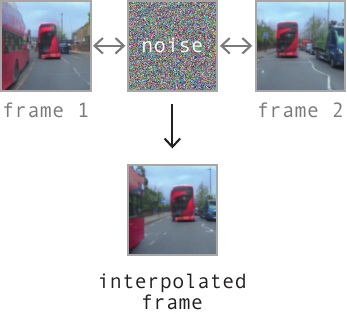}}
\end{tabular}
\caption{Video decoder training tasks. Each task is defined by masking ground truth images and context tokens. We pass noise as input for frames we want to predict. Tokens are provided for predicted frames except for video interpolation task where the diffusion process is guided solely by image context.}
\label{fig:decoder-tasks}
\end{figure}

To train our video diffusion decoder for multiple inference tasks we take inspiration from \citep{harvey2022flexible} where we can perform multiple tasks by masking certain frames or the conditioning image tokens. We choose to train a single video diffusion model for all tasks as it has been shown that multi-task training improves performance on individual tasks \citep{harvey2022flexible}. The tasks include image generation, video generation, autoregressive decoding, and video interpolation. Each task is sampled equally. For example, for the autoregressive generation task, we provide previously generated past frames as context and conditioning image tokens for frames we want to predict. We include both forward and backward autoregressive tasks. See \Cref{fig:decoder-tasks} for examples of each task. 
We also apply a conditioning dropout by randomly masking out each conditioning image token with probability $p=0.15$ as it helps the model generalize beyond relying on tokens for information and improves temporal consistency.

The video decoder is trained on the noise prediction objective. More specifically, we use the $\rvv$-parameterization as proposed in \citep{salimans22} because it avoided unnatural color shifts and maintained long-term consistency as similarly found in \citep{ho2022imagen}. In practice, we use a weighted average of $L_1$ and $L_2$ losses. The video decoder loss $L_{\text{video}}$ is:

\begin{equation}
\label{eq:diffusion-loss}
L_{\text{video}}= \mathbb{E}_{\mathbf{\epsilon}, t'} \left[\lVert 
\mathbf{\epsilon}_{\theta}(\rvx^{t'}, t', \rvz, \rvm)
-
\mathbf{\epsilon}
\rVert^2_2 \right]
\end{equation}

where:
\begin{itemize}
\item $\mathbf{\epsilon}_\mathbf{\theta}$ is the denoising video model.
\item $\epsilon$ is the denoising target, which uses the $\rvv$-parameterization.
\item $t' \sim U(0,1)$ is the sampled discrete diffusion time.
\item $\rvx=(\rvx_1, ..., \rvx_{T'})$ is a video sequence of length $T'$.
\item $\rvx^{t'} = \alpha_{t'}\mathbf{x} + \sigma_{t'}\mathbf{\epsilon}$ represents the noised video, with $\alpha_{t'}$ and $\sigma_{t'}$ functions of $t'$ that define the noise schedule.
\item $\rvz=(\rvz_1, ..., \rvz_{T'}) = E_{\theta}(\rvx)$ is the sequence of conditioning image tokens.
\item $\rvm=(\rvm_1, ..., \rvm_{T'})$ is a sequence of image masks as specified by the training task (see \Cref{fig:decoder-tasks}).
\end{itemize}


\section{Data}
\label{section:data}

\begin{figure}[t]
    \centering 
\begin{subfigure}{0.32\textwidth}
  \includegraphics[width=\linewidth]{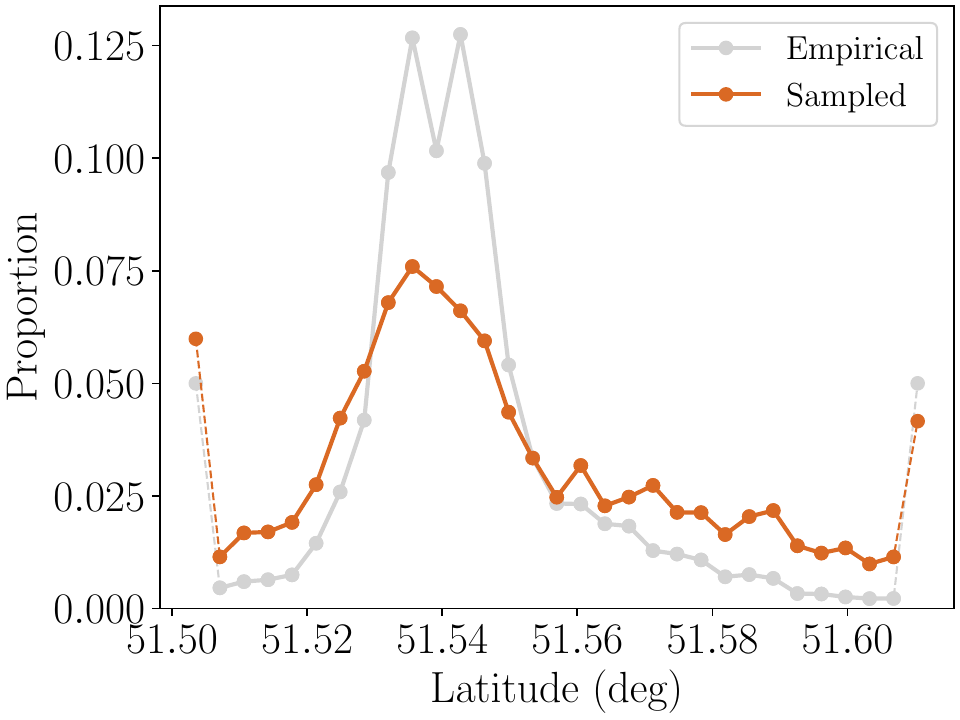}
  \label{fig:data_sampling_1}
\end{subfigure}\hfil 
\begin{subfigure}{0.32\textwidth}
  \includegraphics[width=\linewidth]{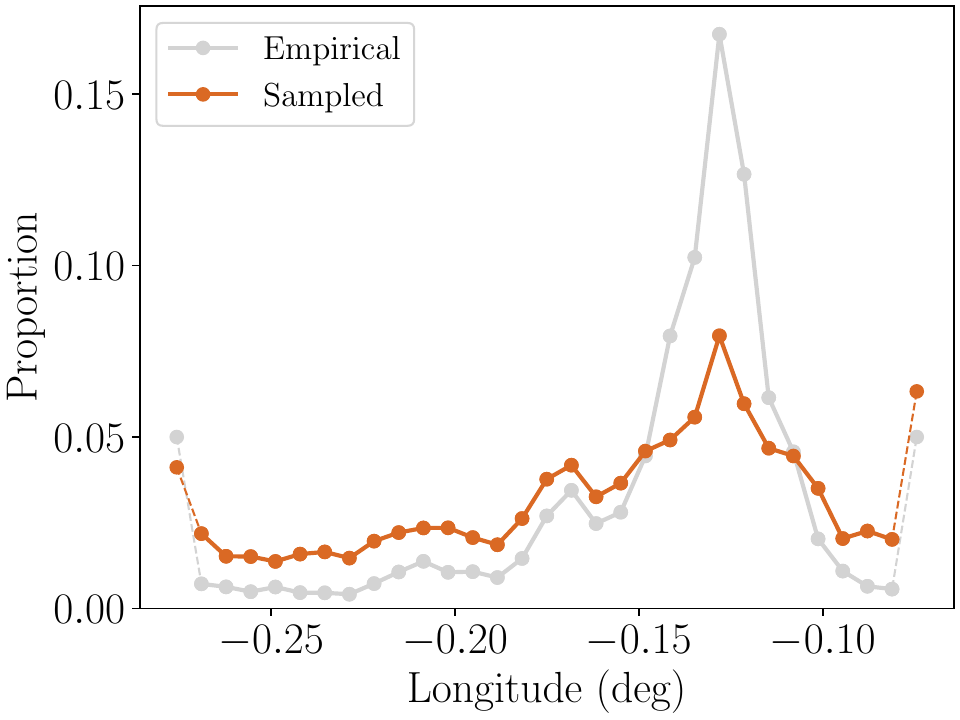}
  \label{fig:data_sampling_2}
\end{subfigure}\hfil
\begin{subfigure}{0.32\textwidth}
  \includegraphics[width=\linewidth]{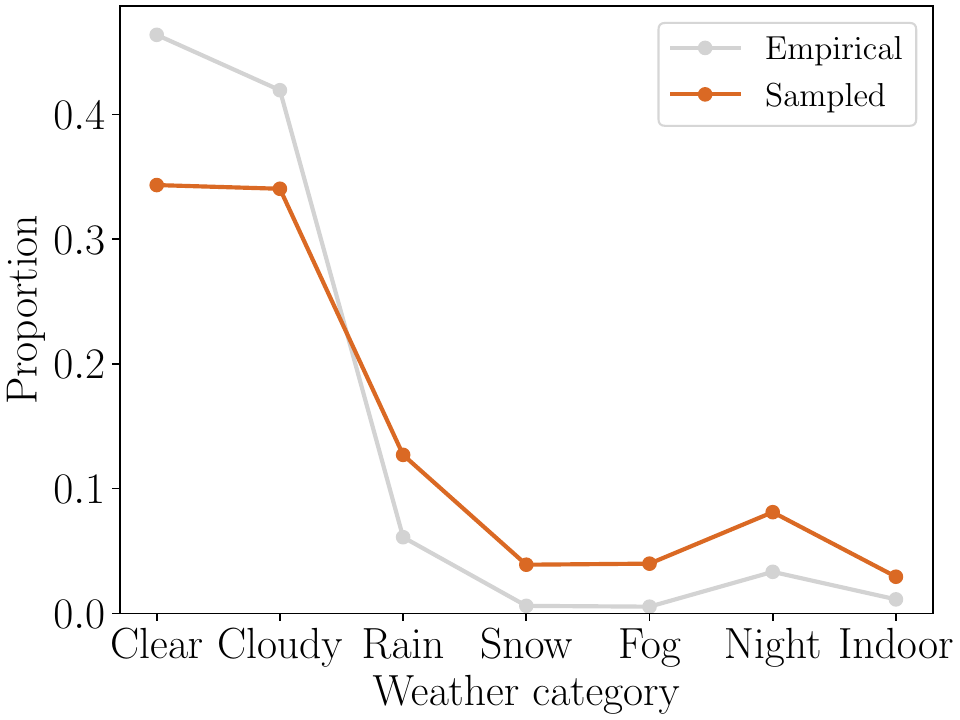}
  \label{fig:data_sampling_3}
\end{subfigure}

\begin{subfigure}{0.33\textwidth}
  \includegraphics[width=\linewidth]{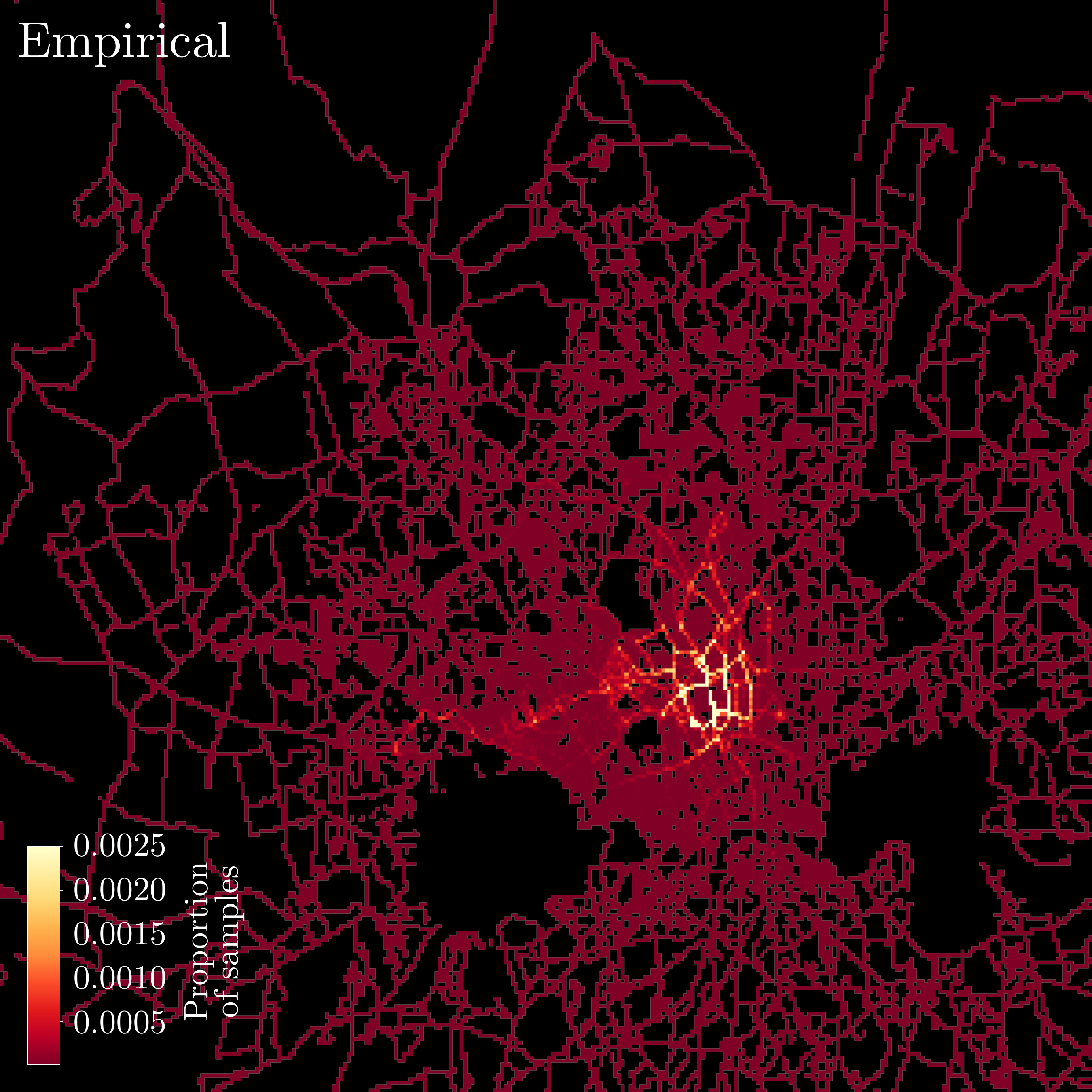}
  \label{fig:data_sampling_4}
\end{subfigure}\hfil 
\begin{subfigure}{0.33\textwidth}
  \includegraphics[width=\linewidth]{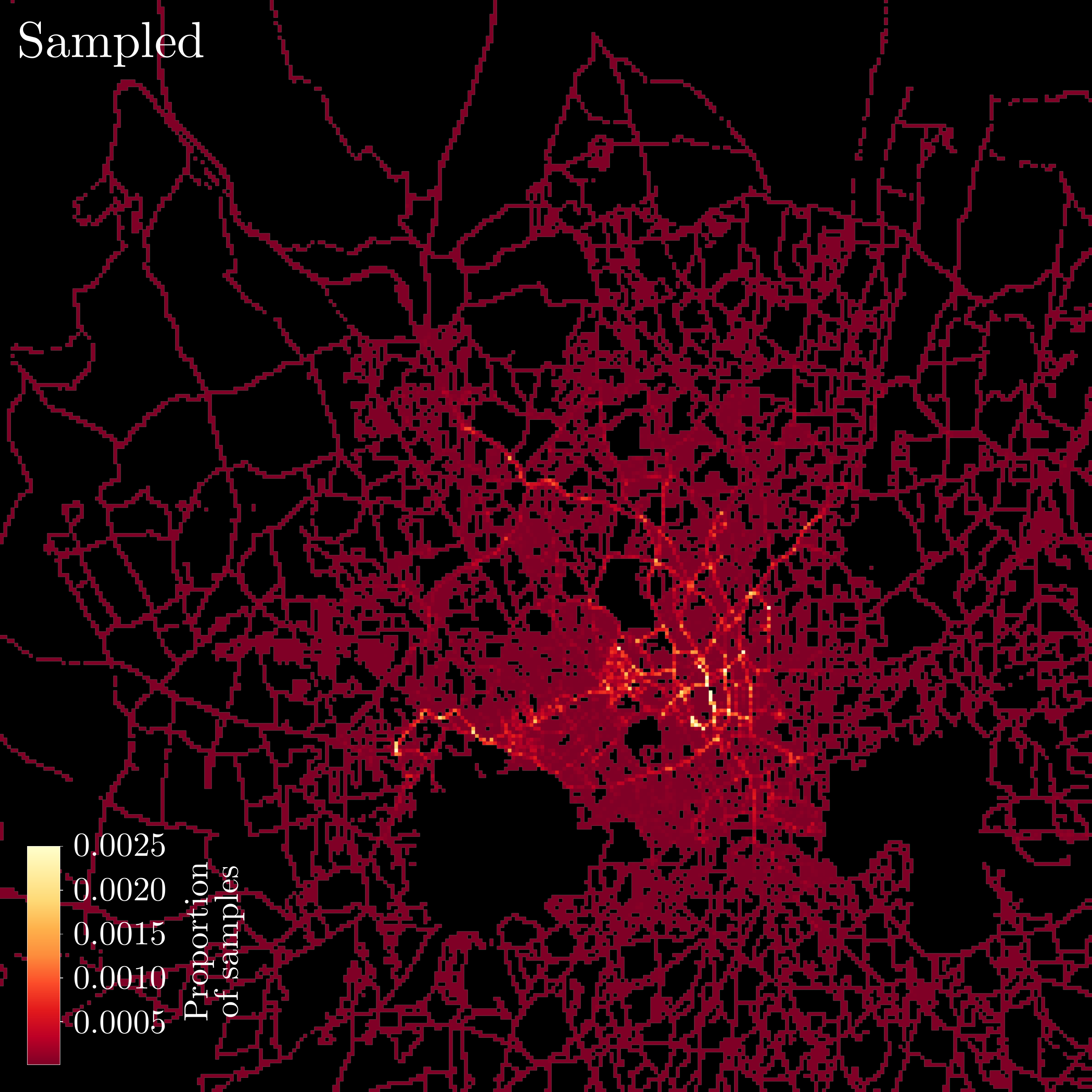}
  \label{fig:data_sampling_5}
\end{subfigure}\hfil 
\begin{subfigure}{0.33\textwidth}
  \includegraphics[width=\linewidth]{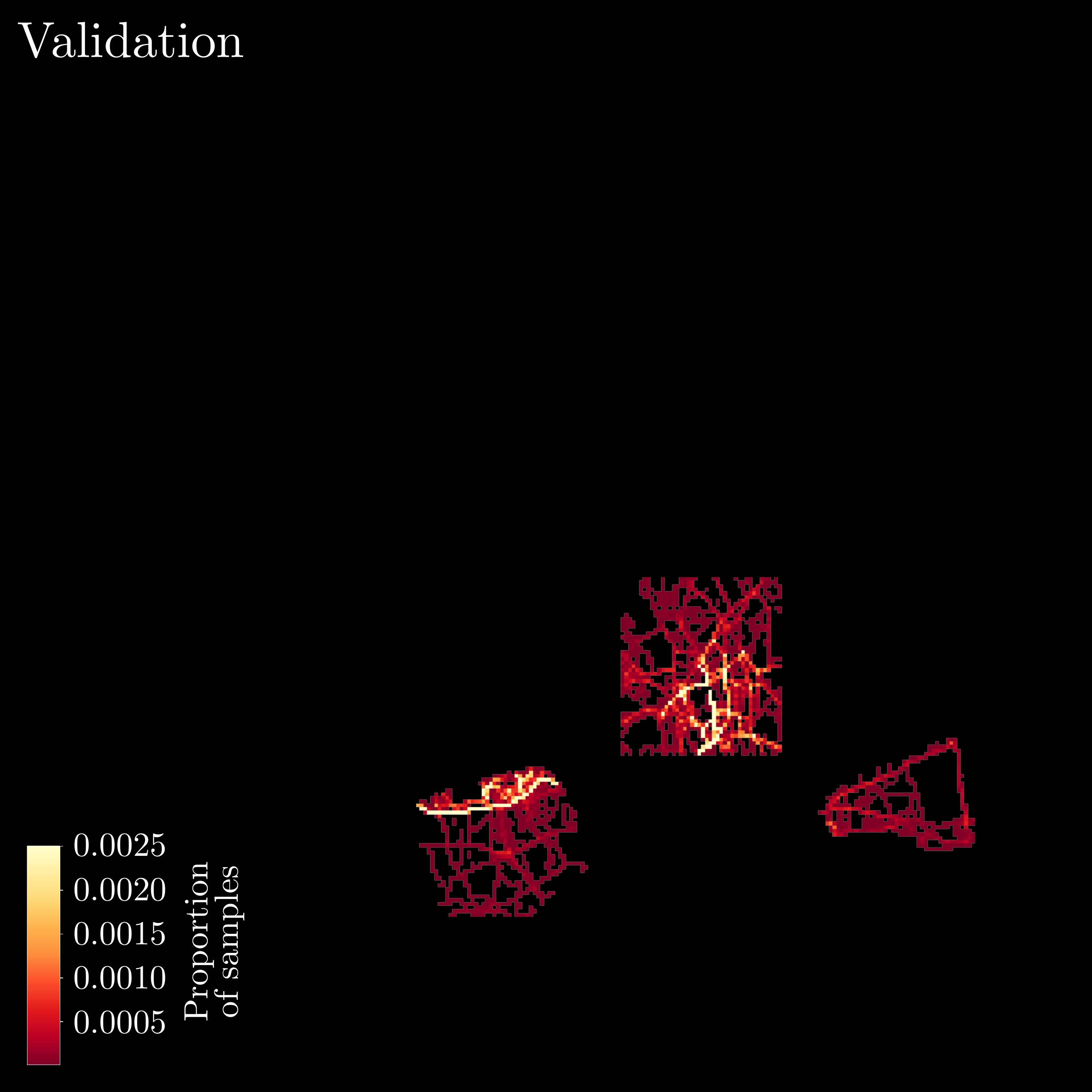}
  \label{fig:data_sampling_6}
\end{subfigure}
\caption{Data sampling. The top row shows the empirical distribution and the sampled distribution for three features we selected to balance over during training: latitude, longitude and weather condition. Dashed lines indicate data outside of the range we balanced over. The bottom row shows the geographical heatmap of sampled latitude and longitude coordinates for the whole training set, the sampled training set and the geofenced validation set.}
\label{fig:data_sampling}
\end{figure}

Our training dataset consists of 4,700 hours at 25Hz of proprietary driving data collected in London, UK between 2019 and 2023. This corresponds to approximately 420M unique images. During training we balance over a customizable set of features to control the distribution of data (\Cref{fig:data_sampling}). We achieve this by sampling individual data points with weighting inversely proportional to the (binned and precomputed) empirical distribution of a given feature. For a given example we take the joint probability across all features to balance and stochastically decide whether to include or discard that example. We can control the strength of balancing by raising the sampling weight to an exponent, where an exponent of 0 would result in the empirical distribution (no balancing) and an exponent of 1 would result in a uniformly balanced distribution. We used an exponent of 0.5 for all features as a compromise between final balancing achieved and the severity of discarding samples for training efficiency.

For the tokenizer we balanced over (latitude, longitude, weather category) to account for geography and visually distinct weather conditions ensuring our tokenizer can adequately represent a diverse range of scenes.

For the world model and the video diffusion model we balanced over (latitude, longitude, weather category, steering behavior category, speed behavior category), additionally considering speed and steering behaviors to ensure the dynamics of different behaviors are captured and sufficiently modeled by the world model and the temporal decoder.

Our validation dataset contains 400 hours of driving data from runs not included in the training set. The runs selected for validation are those that pass through predetermined geofences as well as a selection of randomly selected runs. We further split our validation set into strict geofences in order to analyze only those samples strictly within the validation geofence (i.e., roads never seen during training) and another geofence around our main data collection routes (i.e., roads seen during training) as a way to monitor overfitting and generalization.

\section{Training Procedure}
\label{section:training-procedure}

    
In this section, we describe how the three trainable components of GAIA-1 were optimized. We provide details of hyperparameter configurations, hardware used and training times.

\subsection{Image Tokenizer} 
The image tokenizer (0.3B parameters) was trained on images of resolution $H\times W = 288 \times 512$ ($9/16$ ratio). The spatial downsampling of the encoder is $D=16$, therefore each image is encoded as $n=18\times 32 = 576$ discrete tokens with a vocabulary size $K=8192$. The bit compression is $\frac{288\times512\times3\times8}{18\times32\times 13} \approx 470$. 

The discrete autoencoder was optimised with AdamW \citep{loshchilov19} and a learning rate of $1\times 10^{-4}$, weight decay $0.01$, beta coefficients $(0.5, 0.9)$. The loss weights are $\lambda_{L_1}=0.2$, $\lambda_{L_2}=2.0$, $\lambda_{L_{\text{perceptual}}}=0.1$, $\lambda_{L_{\text{GAN}}}=1.0$, $\lambda_{L_{\text{codebook}}}=1.0$, $\lambda_{L_{\text{DINO}}}=0.1$.

The model was trained for 200k steps in 4 days with a batch size equal to 160, split across 32 A100 80GB GPUs. We used 5k of linear warm-up and 10k of cosine decay to a final learning rate of $1\times 10^{-5}$.

\subsection{World Model}
The world model (6.5B parameters) was trained on video sequences of size $T = 26$ at \SI{6.25}{\hertz}, which correspond to 4s-long videos. The text was encoded as $m=32$ text tokens per time step, and the action as $l=2$ tokens. The total sequence length of the world model is therefore $T \times (m + n + l) = 15860$. 

The world model was optimized with AdamW and a learning rate of $1\times 10^{-4}$, weight decay $0.1$, beta coefficients $(0.9, 0.95)$, norm gradient clipping $1.0$. Training examples were either unconditioned, action-conditioned, or text conditioned. The ratios of these respective conditioning modes were 20\%/40\%/40\%.

The model was trained for 100k steps in 15 days, with 2.5k of linear warm-up and 97.5k of cosine decay reducing the learning rate by a factor of 10 over the course of training. The batch size was 128 split across 64 A100 80GB GPUs. We used the FlashAttention v2 implementation~\citep{dao2022flashattention} in the transformer module, as it offered significant advantages in terms of both memory utilization and inference speed. To optimize distributed training, we used the Deepspeed ZeRO-2 training strategy \citep{rasley20} with activation checkpointing.

\subsection{Video Decoder}

The video decoder (2.6B) was trained on sequences of $T'=7$ images of resolution $H\times W = 288 \times 512$ sampled from the dataset at either \SI{6.25}{\hertz}, \SI{12.5}{\hertz} or \SI{25}{\hertz}. The training tasks (\Cref{fig:decoder-tasks}) were sampled with equal probability. We used a cosine $\beta$-noise schedule \citep{hoogeboom2023simple}.

The video decoder was optimized with AdamW and a learning rate of $5\times 10^{-5}$, weight decay $0.01$, beta coefficients $(0.9, 0.99)$, norm gradient clipping $1.0$. The model was trained for 300k steps in 15 days, with 2.5k of linear warm-up and 5k of cosine decay to a final learning rate of $1\times 10^{-6}$. We used a weighted average of $L_1$ and $L_2$ losses with weights $\lambda_{L_1}=0.1$ and $\lambda_{L_2}=1.0$. The batch size was 64 split across 32 A100 80GB GPUs. We used an exponential moving average for the parameters with a decay of $0.999$. The training strategy was also Deepspeed ZeRO-2 with activation checkpointing.

\section{Inference}
\label{section:inference}
In this section, we describe in more detail the inference procedure of the world model and the video decoder.

\subsection{World Model}

\begin{figure}[t]
    \centering 
    \begin{subfigure}{0.32\textwidth}
      \includegraphics[width=\linewidth]{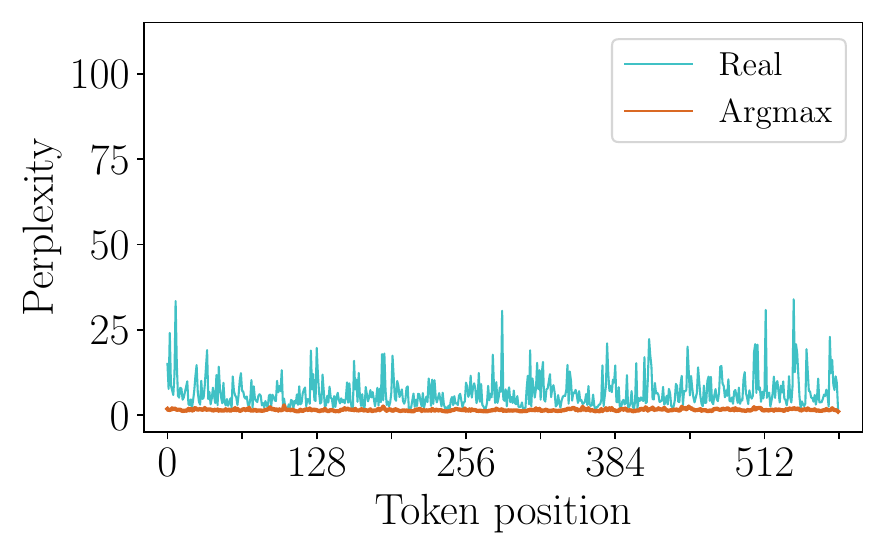}
      \caption{Argmax.}
    \end{subfigure}\hfill
    \begin{subfigure}{0.32\textwidth}
      \includegraphics[width=\linewidth]{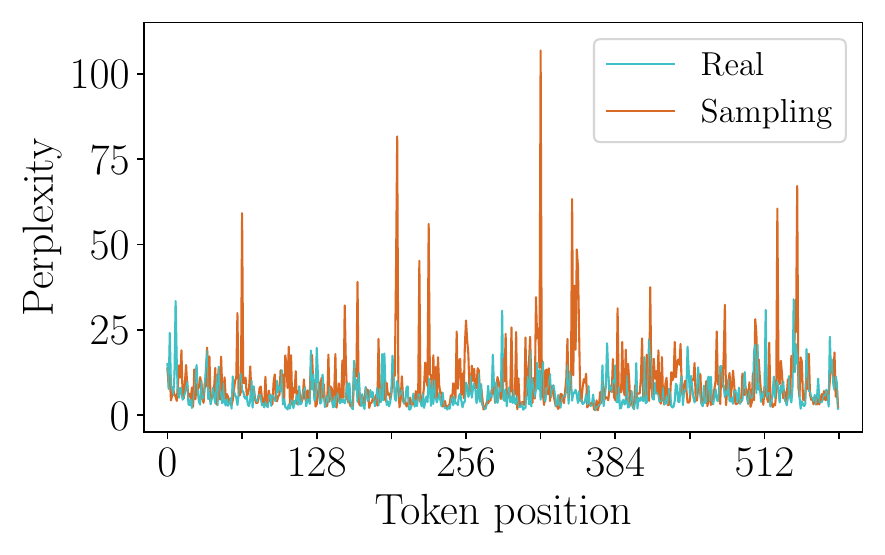}
      \caption{Sampling.}
    \end{subfigure}\hfill
    \begin{subfigure}{0.32\textwidth}
      \includegraphics[width=\linewidth]{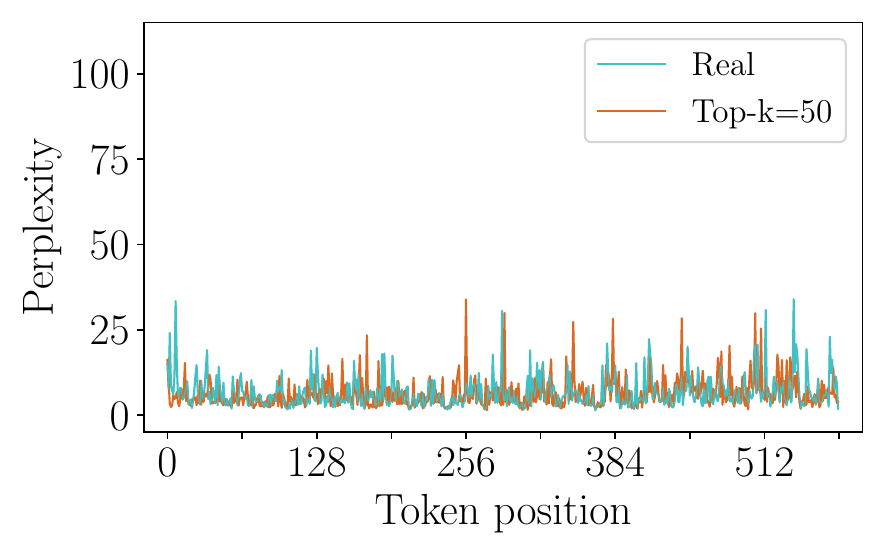}
      \caption{Top-k sampling.}
    \end{subfigure}
\caption{Perplexity of the world model as a function of the position of the generated token. We consider the $n=576$ tokens of a single image frame. We compare the perplexity of the tokens from a real image, to those generated with the following strategies: argmax, sampling, or top-k. In (a), by inspecting the perplexity of a real image we notice it oscillates between low and high values, meaning there is a good range of diversity in those tokens. In contrast, if we look at the argmax strategy, we notice the perplexity only takes extremely low values (no diversity, manifesting in the predicted frames to repeat themselves). Conversely in (b), if we sample from the entire distribution, the perplexity of some tokens can take extremely high values, due to sampling from the unreliable tail. In (c), we observe that top-k=50 sampling produces tokens that have a similar perplexity distribution to real tokens.}
\label{figure:perplexity}
\end{figure}

\paragraph{Sampling.} The world model autoregressively predicts the next image token, conditioned on previous text, image and action tokens. Given the past tokens we perform $n$ forward steps to generate one new image frame. At each step we must sample a token from the predicted logits to select the next token in our sequence. Empirically we observed that maximization-based sampling (i.e. argmax) generates futures that get stuck in a repetitive loop, similarly to language models \citep{holtzman20}. Conversely, if we simply sample from the logits, the selected token can come from the unreliable tail of the probability distribution, which throws the model out-of-distribution, see \Cref{figure:perplexity}.

To encourage diversity as well as realism we employ top-k sampling to sample the next image token from the top-k most likely choices. The chosen value of k is a function of the number of tokens that constitute an image frame as well as the pre-learnt codebook (vocabulary) size.

Our world model can be used to roll out possible futures given starting context as well as generating futures from scratch without any starting context. For long video generation, if the length of the video exceeds the context length of the world model, we employ a sliding window.

\paragraph{Text-conditioning.}The video prediction can be prompted, and thus directed, with text. At training time, we condition our video sequences with text coming from either online narration or offline metadata sources. Because these text sources are imperfect, to improve the alignment between generated futures and the text prompt, we employ classifier-free guidance \citep{ho2022classifier, chang2023muse} at inference time. The effect of guidance is to increase text-image alignment by decreasing the diversity of possible samples. More precisely, for each next token to predict, we compute logits conditioned on text as well as logits with no conditioning (unconditioned). At inference, we can then amplify the differences between the unconditioned and the text-conditioned logits with a scale factor to give the final logits used for sampling.
\begin{equation}
\label{eq:classifier-free-guidance}
l_{\text{final}} = (1 + t)l_{\text{conditioned}} - tl_{\text{unconditioned}}
\end{equation}

By substituting the unconditioned logits with those conditioned on another text prompt, we can perform ``negative'' prompting \citep{negprompt2022}. Pushing the logits away from the negative prompt and towards the positive one encourages the future tokens to include the ``positive'' prompt features while removing the ``negative'' ones.

\begin{figure}[t]
    \centering 
	\begin{subfigure}{0.74\textwidth}
		\includegraphics[width=\linewidth]{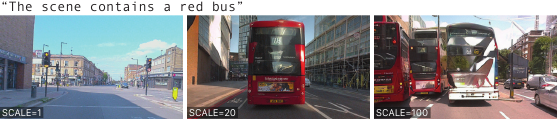}
		\caption{Guidance scale factor}
	\end{subfigure}\hfil 
	\begin{subfigure}{0.24\textwidth}
		\includegraphics[width=\linewidth]{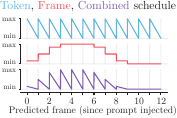}
		\caption{Guidance schedule}
	\end{subfigure}
\caption{Classifier-free guidance.}
\label{fig:guidance-schedule}
\end{figure}

We found it was important to schedule the scale factor used for guidance over tokens as well as frames. Scheduling over tokens allows some to be sampled with high guidance (hence adhering strongly to the prompt) and others to be sampled with low guidance (hence increasing sample diversity). Scheduling over frames allows for controlling the transition from earlier frames as well as mitigating compounding guidance over subsequent consecutive frames. In \Cref{fig:guidance-schedule} we show an example guidance schedule over twelve frames. Typically we used a schedule that sampled tokens with linearly decreasing guidance over tokens and we lowered the guidance over future frames with a cosine decay, with or without an initial plateau. We note that guidance scale and schedule are hyperparameters to be tuned to particular use cases. 

\subsection{Video Decoder}
To decode a sequence of generated tokens from the world model, we use the following video decoding method:

\begin{enumerate}
  \item Decode the first $T'=7$ frames, conditioned on the corresponding $T'$ image tokens.
  \item Autoregressively decode the next $T'-2$ frames, using 2 past overlapping frames as image context, and the following $T'-2$ image tokens.
  \item Repeat the autoregressive process until the $N$ frames have been generated at \SI{6.25}{\hertz}.
  \item Temporally upsample the $N$ frames from \SI{6.25}{\hertz} to \SI{12.5}{\hertz}
  \item Temporally upsample the $2N-1$ frames from \SI{12.5}{\hertz} to \SI{25.0}{\hertz}
\end{enumerate}

We use the DDIM sampler \citep{song2021denoising} with $50$ diffusion steps. During autoregressive decoding, we see a trade-off between reflecting token information content in the generated video and temporal consistency. To balance between these two objectives, we calculate a weighted average of the two tasks \citep{esser2023structure}.
\begin{equation}
\Tilde{\mathbf{\epsilon}}_{\theta}(\rvx^{t'}, t', \rvz, \rvm)
= w\cdot \mathbf{\epsilon}_{\theta}^{\pi}(\rvx^{t'}, t', \rvz, \rvm) + (1-w)\cdot \mathbf{\epsilon}_{\theta}(\mathbf{x}^{t'}, t', \mathbf{z}, \rvm)
\end{equation}

where function $\mathbf{\epsilon}_{\theta}^{\pi}(\rvx^{t'}, t', \rvz, \rvm)$ denoises each frame individually  as images and function $\mathbf{\epsilon}_{\theta}(\mathbf{x}^{t'}, t', \mathbf{z}, \rvm)$ denoises the sequence of frames jointly as a video. In practice, we simply switch on and off the temporal layers. We apply this weighted average randomly for each diffusion step with probability $p=0.25$ and weight $w=0.5$.

While exploring different inference approaches for video decoding we found that decoding video frames autoregressively backwards starting from the end of the sequence led to more stable objects and less flickering on the horizon. In our overall video decoding method, we thus decode the last $T'$ frames and autoregressively decodes the remaining frames backward from there.


\section{Scaling}
\label{section:scaling}

\begin{figure}[t]
    \centering
    \begin{subfigure}[b]{0.49\textwidth}
    \centering
    \includegraphics[width=\linewidth]{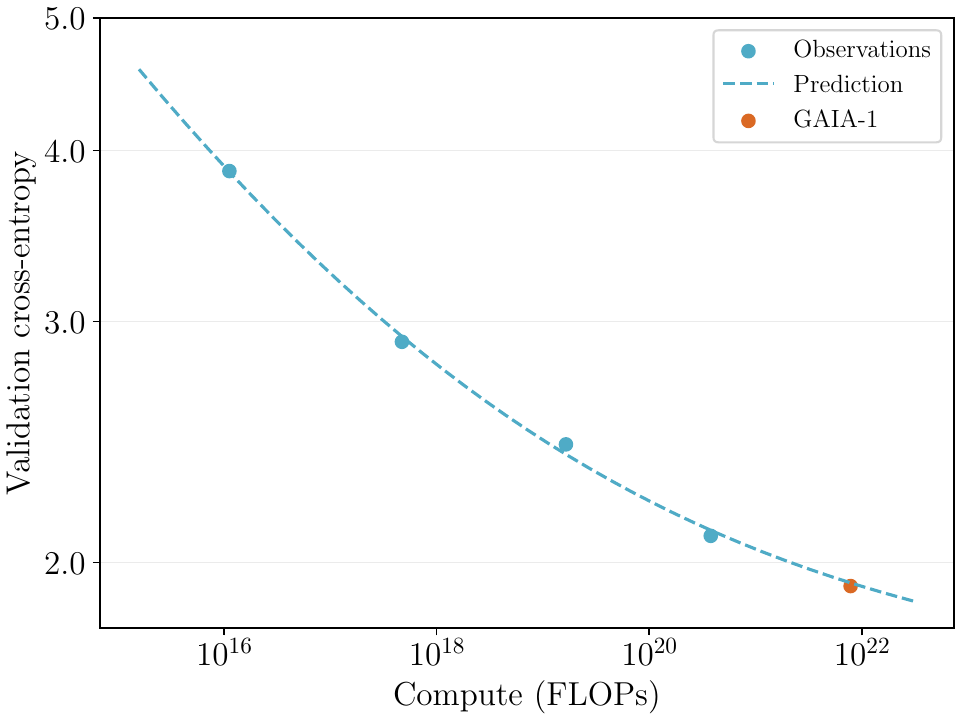}%
    \caption{The final performance of the GAIA-1 world model could be predicted with smaller models trained with less than $20\times$ the compute.}
    \label{figure:predict-performance}
    \end{subfigure}\hfill
    \begin{subfigure}[b]{0.49\textwidth}
    \centering
    \includegraphics[width=\linewidth]{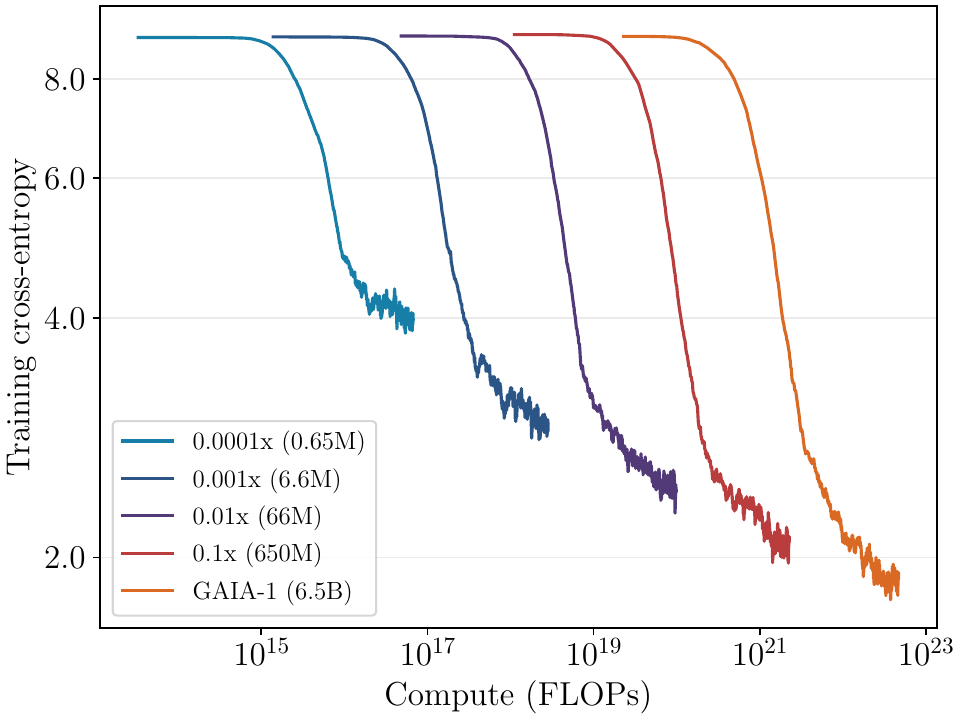}%
    \caption{Training loss curves for world models up to 10,000x smaller. We used an exponential moving average to smooth the training loss curves.}
    \label{figure:compute}
    \end{subfigure}
    \caption{Validation and training cross-entropy of the world model.}
\end{figure}

The formulation of the world modeling task in GAIA-1 shares a commonality with the approach frequently used in large language models (LLMs). In both instances, the task is streamlined to focus on predicting the next token. Although this approach is adapted for world modeling in GAIA-1 rather than the traditional language tasks seen in LLMs, it is intriguing to observe that scaling laws \citep{kaplan20, Chowdhery2022PaLMSL,hoffmann22}, analogous to those observed in LLMs, are also applicable to GAIA-1. This suggests the broader applicability of scaling principles in modern AI models across diverse domains, including autonomous driving.


To explore scaling laws with GAIA-1, we predicted the final performance of the world model using models trained with less than $20\times$ the compute. We evaluated those models on a held-out geofenced validation set by measuring cross-entropy. A power-law of the form $f(x) = c + (x/a)^b$ was then fitted to the data points. In \Cref{figure:predict-performance} we can see that the final cross-entropy of GAIA-1 could be predicted with high accuracy.

The models used to fit the power-law ranged from 10,000x to 10x smaller models in terms of parameters (0.65M to 650M), as visualized in \Cref{figure:compute}. Similarly to \citep{kaplan20}, the compute was estimated as a function of the parameter count. If we denote by $C$ the compute and by $N$ the parameter count (excluding embedding layers), the number of floating point operations for a forward-backward pass of a single token is given by $C=6N$. To obtain the total amount of compute, this value is multiplied by the number of training tokens.

It is worth noting that our extrapolation leads us to the conclusion that there is substantial potential for further improvement through the expansion of both data and computational resources.

\begin{figure}[h!]
    \centering
    \includegraphics[width=\linewidth]{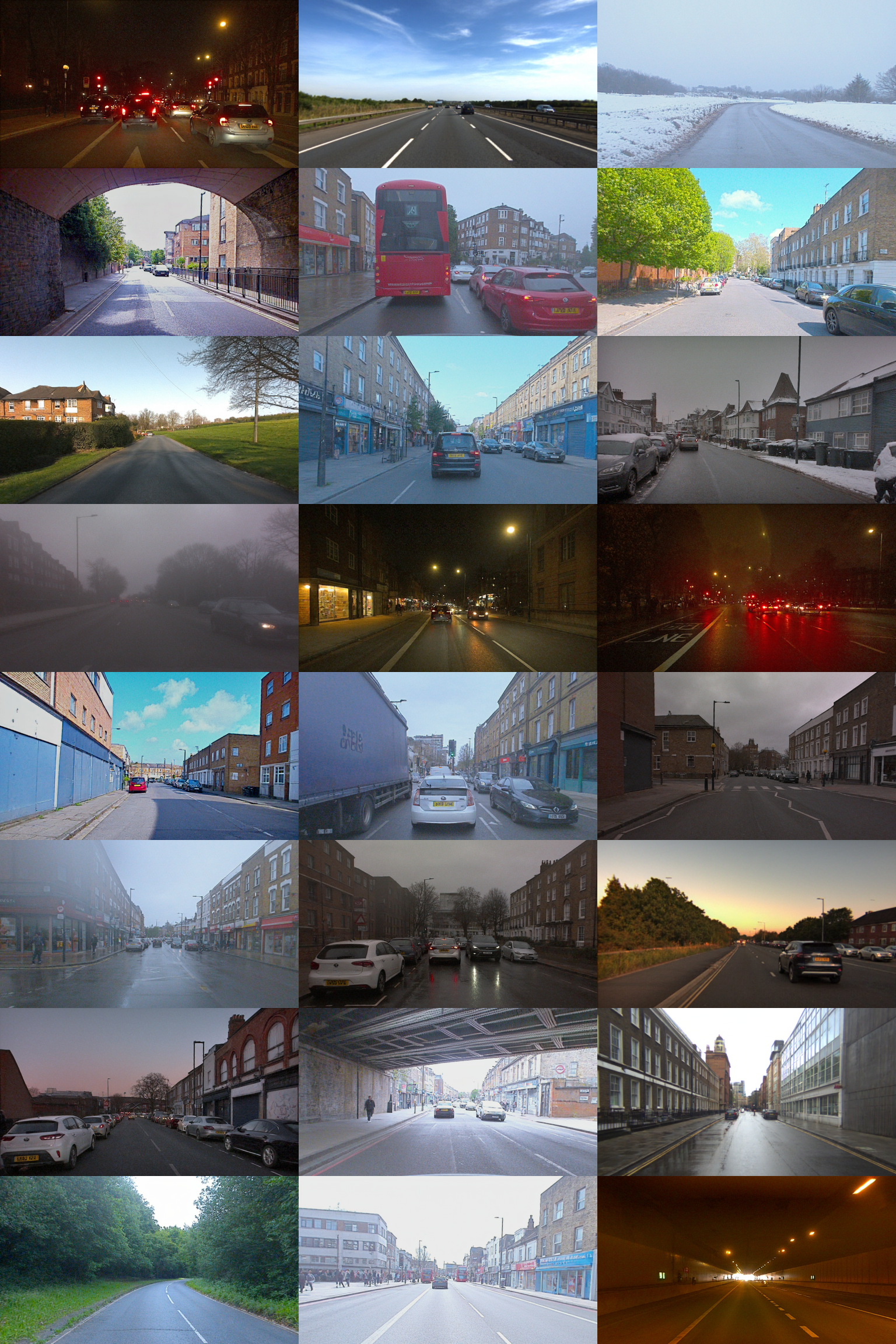}%
    \caption{Images generated by GAIA-1, highlighting the diversity of the generated driving scenes.}
    \label{fig:tile_montage}
\end{figure}

\section{Capabilities and Emerging Properties}
\label{section:emerging-properties}

In this section we showcase the capabilities and emerging properties of GAIA-1 through a series of qualitative examples. The comprehensive list of video examples can be found \href{https://www.youtube.com/playlist?list=PL5ksjZd5b6SI-6MQi6ghoD-GilTPmsQIf}{here}. Figure~\ref{fig:tile_montage} shows the variety of scenarios that can be generated by our model. As evidenced by the examples presented in the rest of this section, GAIA-1 exhibits a level of understanding and summarization of the generative rules of the world through the following emergent properties:
\begin{enumerate}
    \item Learning high-level structures and scene dynamics: it generates coherent scenes with objects positioned in plausible locations and exhibiting realistic object interactions, such as traffic lights, rules of the road, give ways, etc. This suggests that the model is not just memorizing statistical patterns but is understanding the underlying rules that govern the arrangement and behavior of objects in the world (see \Cref{subsection:long_generation}).
    \item Generalization and creativity: it can generate novel and diverse videos that go beyond specific instances in the training set. It can produce unique combinations of objects, movements, and scenes that were not explicitly present in the training data, demonstrating remarkable extrapolation capabilities. This demonstrates a certain level of generalization and creativity, which suggests an understanding of the underlying generative rules that govern video sequences (see \Cref{subsection:multimodal}). 
    \item Contextual awareness: GAIA-1 can capture contextual information and generate videos that reflect this understanding. For example, it can generate coherent actions and responses in videos based on the initial conditions or the context provided. Moreover, GAIA-1 exhibits the understanding of 3D geometry, effectively capturing the intricate interplay of pitch and roll induced by road irregularities (e.g. speed bumps). This contextual awareness suggests that the models are not merely reproducing statistical patterns but are actively processing and summarizing the given information to generate appropriate video sequences (see \Cref{subsection:finegrained}).
\end{enumerate}

\subsection{Generation of Long Driving Scenarios}
\label{subsection:long_generation}
\begin{figure}[t]
    \centering
    \includegraphics[width=\linewidth]{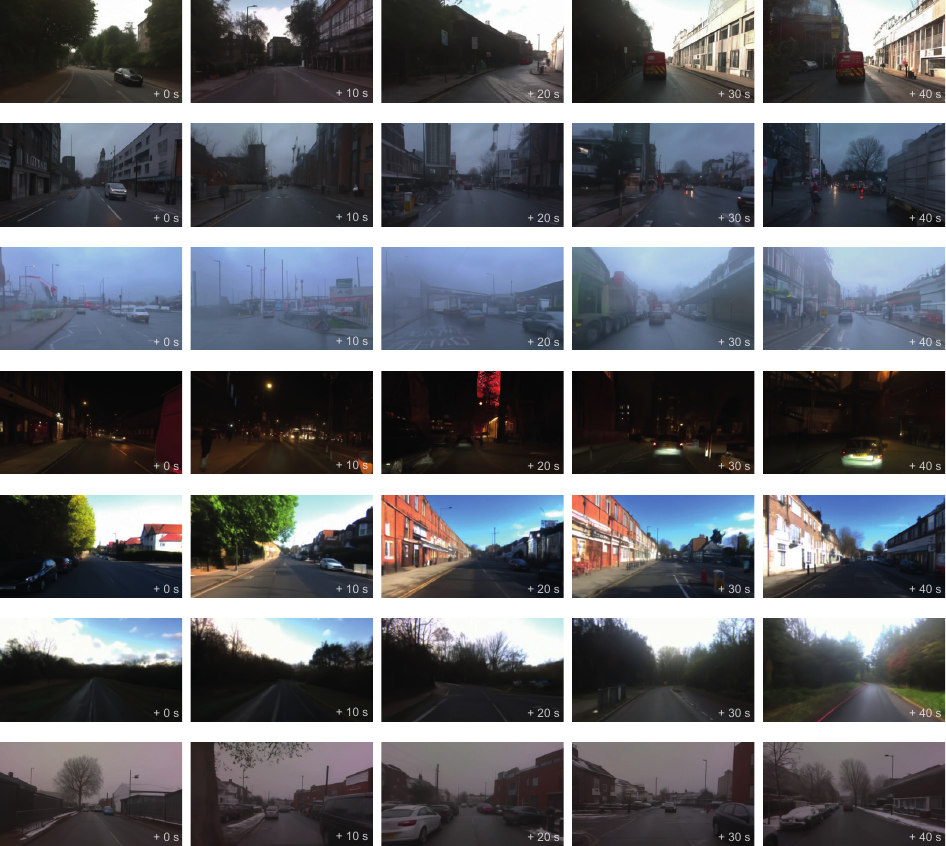}%
    \caption{Long, diverse driving scenarios generated entirely from imagination by the world model.}
    \label{fig:emergent_long}
\end{figure}

GAIA-1 can generate stable long videos (minutes) entirely from imagination (\Cref{fig:emergent_long}). In order to do this, the model leverages its learned implicit prior distribution of the world to generate fully-imagined realistic driving scenarios, with complex road layouts, buildings, cars, pedestrians, and more. This is a demonstration that GAIA-1 understands the rules that underpin the world we inhabit and its structures and dynamics.

\subsection{Generation of Multiple Plausible Futures}
\label{subsection:multimodal}
\begin{figure}[t]
    \centering
    \includegraphics[width=\linewidth]{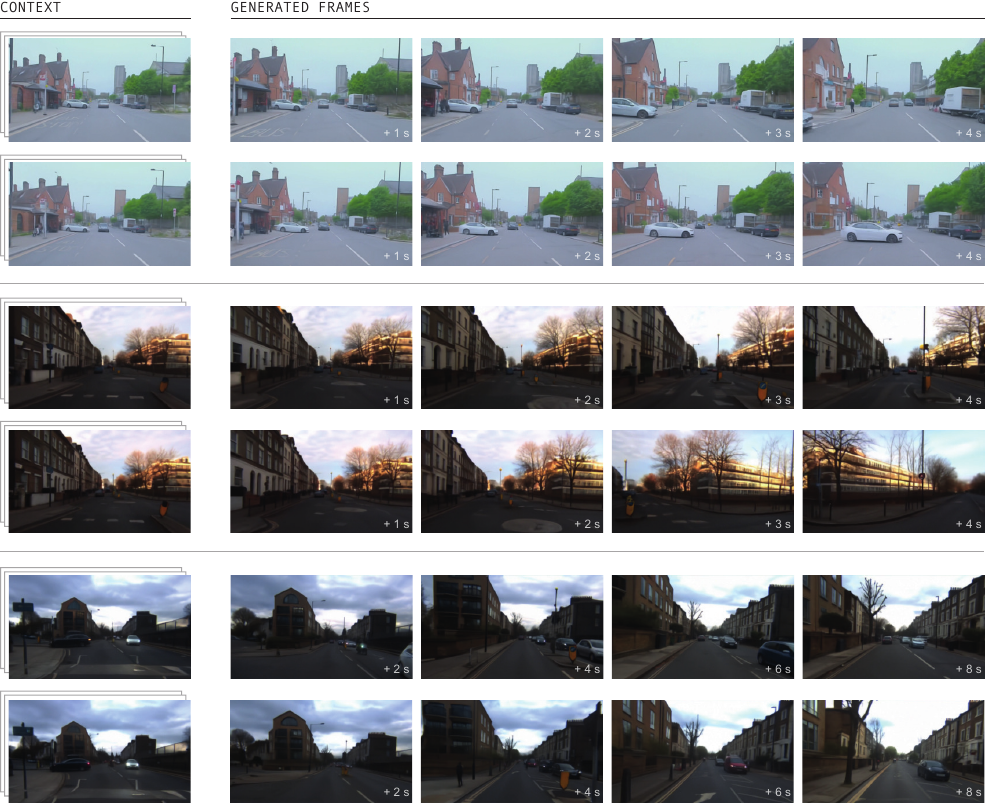}%
    \caption{Examples of multiple plausible futures predicted by the world model from a given video context. 1) We observe a complex giving way interaction between the white vehicle and the ego-vehicle. In the first future, the white vehicle reverses to give way to the ego-vehicle. In the second future, the opposite occurs and the ego-vehicle slows down to give way to the white vehicle. 2) We see two plausible ego-behaviors: going straight or turning right at the roundabout. 3) The model predicts two futures with varying traffic levels.}
    \label{fig:emergent_multimodal}
\end{figure}

GAIA-1 has the ability to generate a variety of distinct future scenarios based on a single initial prompt. When presented with a brief video as context, it can generate numerous plausible and diverse outcomes by repeatedly sampling. GAIA-1 accurately models multiple potential future scenarios in response to the video prompt while maintaining consistency with the initial conditions observed in the video. As seen in \Cref{fig:emergent_multimodal}, the world model can reason about (i) dynamic interactions with road users (e.g. giving way or not giving way), (ii) multimodal ego-behaviors (e.g. going straight or turning at a roundabout), and (iii) multimodal dynamic scene (e.g. variable traffic density and types of road users such as pedestrians, cyclists, motorcyclists, vehicles) and static scene (e.g. road layout, buildings, vegetation).

\subsection{Fine-Grained Control of the Ego-Vehicle Behavior and Driving Scenes}
\label{subsection:finegrained}
\begin{figure}[t]
\centering
\begin{subfigure}{\textwidth}
    \includegraphics[width=\linewidth]{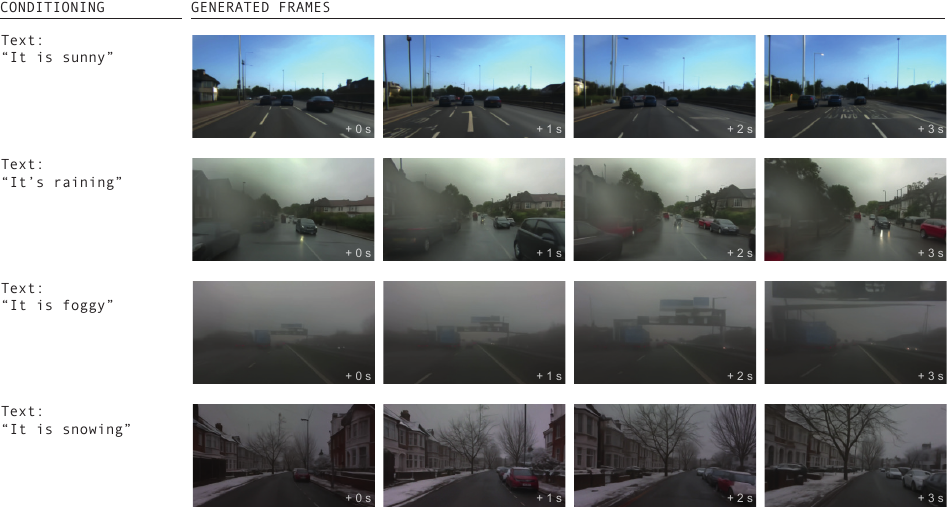}%
    \caption{Weather.}
    \label{fig:emergent_weather}
\end{subfigure}
\par\bigskip
\begin{subfigure}{\textwidth}
    \includegraphics[width=\linewidth]{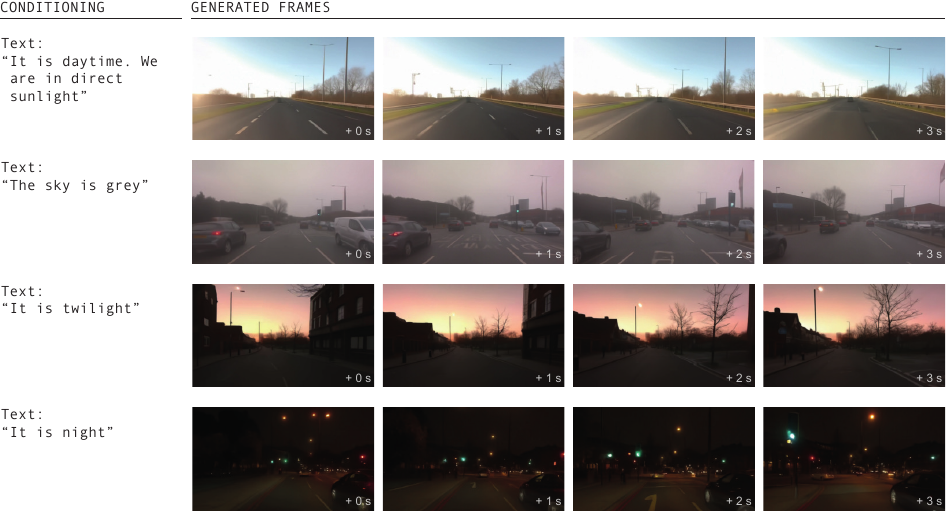}%
    \caption{Illumination.}
    \label{fig:emergent_illumination}
\end{subfigure}
\caption{Generation from a text prompt, showing that the world model has learned different concepts such as weather or illumination.}
\label{fig:text-conditioned}
\end{figure}

GAIA-1 can generate videos from text prompts only, completely imagining the scene. To demonstrate this we showcase how we can generate driving scenarios from text prompts that guide the model towards specific weather or lighting conditions in \Cref{fig:text-conditioned}.

\begin{figure}[t]
    \centering
    \includegraphics[width=\linewidth]{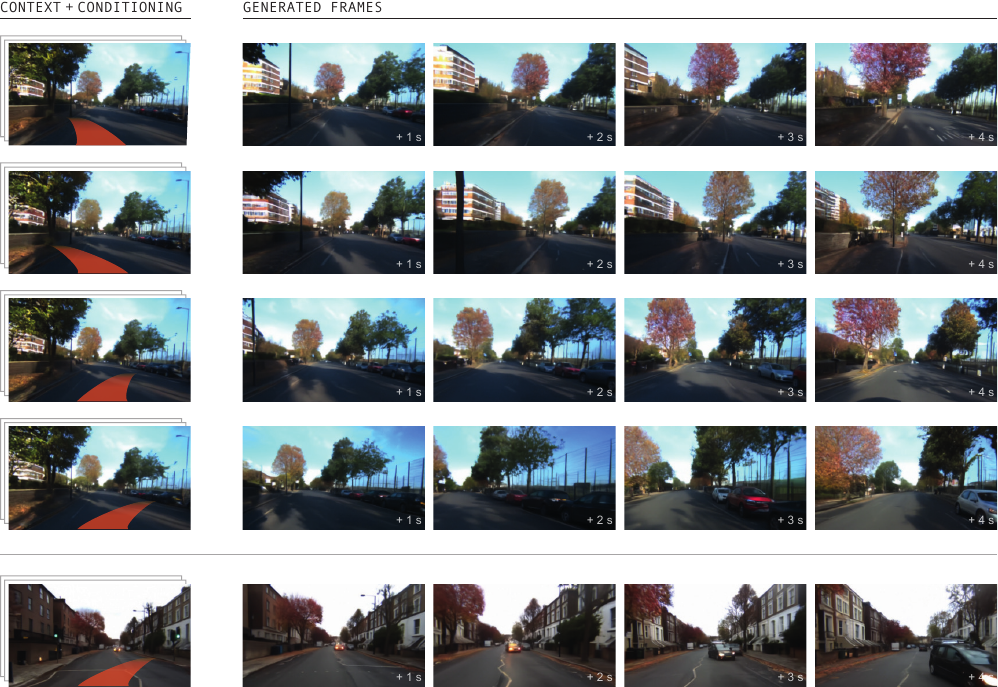}%
    \caption{The world model can predict different outcomes conditioned on its actions. In the first four rows, we execute different out-of-distribution actions (left, strong left, right, strong right --- while maintaining speed) from a given context video. The world model can predict the corresponding states with accurate geometry. In the last row, we force the ego-vehicle to steer right while maintaining speed, and as a consequence, we observe the oncoming vehicle reacting and making a maneuver to avoid a collision.}
    \label{fig:emergent_actions}
\end{figure}

Next, we present compelling examples where the model exhibits fine-grained control over the vehicle dynamics in the video. By leveraging this control, we can prompt the model to generate videos depicting scenarios that lie outside the bounds of the training data. This shows that GAIA-1 is able to disentangle the ego-vehicle dynamics from the surrounding environment and effectively generalize to unfamiliar scenarios. It provides explicit ability to reason about the impact of our actions on the environment (safety), it allows richer understanding of dynamic scenes (intelligence), it unlocks model-based policy learning (planning in the world model), and it enables exploration in closed-loop (by considering the world model as a neural simulator).
To showcase this, we make GAIA-1 generate futures where the ego-vehicle steers left or right, deviating from its lane (\Cref{fig:emergent_actions}). GAIA-1 would never have seen these incorrect behaviors in the expert driving dataset used to train it, indicating that it can extrapolate driving concepts previously unseen in the training data.  We also see realistic reactions of other agents to the ego-vehicle's controlled behavior.

Finally we demonstrate the ability of GAIA-1 to leverage both text and action to fully imagine a driving scenario. In this particular case we prompt the model to generate a bus in front of the ego-vehicle and then we force its actions to overtake the bus (see \Cref{fig:prediction_task}).

\clearpage
\section{Related Work}

\paragraph{Video generative models.} 
Video generative models are neural networks that can generate realistic video samples. They can be grouped in four categories: VAE-based (variational autoencoder \citep{kingma14}, GAN-based (generative adversarial network \citep{goodfellow14}), diffusion-based \citep{sohl-dickstein15}, and autoregressive-based \citep{pixelcnn16}. 

Latent-variable video models (VAE-based) try to infer the underlying latent process that generated the videos \citep{babaeizadeh18,denton18,villegas19,franceschi20,fitvid21}. One known limitation of those models is that they tend to generate blurry outputs due to limited representational power, inadequate choice of prior distribution, and the optimization of a lower-bound instead of the true likelihood. GAN-based methods produce more realistic videos \citep{vondrick16,tulyakov18,clark19,kim2021drivegan,skorokhodov22,brooks22} but are known to suffer from training instability and a lack of generation diversity \citep{goodfellow2016tutorial}. Diffusion-based methods have yielded significant enhancements in realism, controllability, and temporal consistency. They can operate either at the pixel level \citep{ho2022video,harvey2022flexible,voleti22,hoppe22,singer22,molad2023dreamix,ho2022imagen} or in the latent space of a pre-trained image tokenizer \citep{Zhou2022MagicVideoEV,esser2023structure,blattmann2023align}. Diffusion models are expressive neural networks that can fit complex data distributions, but rely on a long Markov chain of diffusion steps to generate samples. Lastly, autoregressive-based methods are conceptually simple and rely on tractable exact likelihood optimization (fits the entire data distribution). Likewise, they can operate at the pixel level \citep{kalchbrenner16,Weissenborn20}, or in a discrete learned token space \citep{yan2021videogpt,moing21,ge2022long,seo22}. A known limitation is the slow generation speed, but this issue could be alleviated by future research on parallel sampling \citep{teco22yan,villegas22phenaki,yu2023magvit}, reducing the number of latent variables \citep{hawthorne22}, and improvements in hardware accelerators.

\paragraph{World models.} 
A world model is a predictive model of the future that learns a general representation of the world in order to understand the consequences of its actions \citep{ha18,lecun22}. The main use cases are: pure representation learning, planning (look-ahead search), or learning a policy in the world model (neural simulator). 

World modeling has been used as a pre-training task to learn a compact and general representation in a self-supervised way \citep{Schwarzer2020DataEfficientRL,wu2023masked}. Subsequently, using this representation as a state for traditional reinforcement learning (RL) algorithms significantly accelerated convergence speed. World models can also be utilized for look-ahead search, in order to plan by imagining the outcomes of future actions. They have proven to be highly effective in game environments or board games \citep{schrittwieser2020mastering,ye2021mastering}. Additionally, world models can be a solution to the sample efficiency issues of RL algorithms by acting as a simulator of the environment \citep{ha18,kaizer19,hafner2021dreamerv2,kim2021drivegan,hafner2023mastering,wu2023daydreamer,wang2023drive}, although this assumes the world model is an accurate model of the environment. 

A recent line of work suggests casting world modeling as a single sequence model, treating states, actions and rewards as simply a stream of data \citep{janner2021sequence,chen2021decisiontransformer,reed22gato,liu2022masked,micheli2023transformers,commavq2023}. The advantage of such a perspective is that world models can benefit from scaling properties of high-capacity sequence model architectures applied to large-scale unsupervised training \citep{brown20}. This is the approach that GAIA-1 takes, leveraging those scaling properties to model  complex environments such as real-world driving scenes.

\paragraph{Scaling.} 
Large language models have shown clear benefits in scaling model size and data \citep{devlin-bert19,2020t5,brown20,megatron22,Chowdhery2022PaLMSL,llama23,openai2023gpt4}. In particular, \citep{kaplan20} showed predictable relationships between model/data size and loss over multiple orders of magnitude. 
\citep{kaplan20} derived power laws for transformer based language models in order to optimally allocate the compute budget between the model and data size. Those laws were then refined by \citep{hoffmann22} by adapting the learning rate schedule when changing the dataset size. Another direction of research to improve the training efficiency of language models is data quality. \citep{du22} showed that the quality of the training data plays a critical role in the performance of language models in downstream tasks. 

Transferring the scaling principles from large language models to the visual domain holds the potential for delivering consistent and expected performance improvements \citep{tay2022scale,dehghani2023scaling,hoogeboom2023simple,ho2022imagen,peebles2023scalable}. In this work, by casting the problem of world modeling as unsupervised sequence modeling, we have shown that similar scaling trends from language models also applied to world models.

\section{Conclusions and Future Work}
\label{section:conclusion}
GAIA-1 is a generative world model for autonomous driving. The world model uses vector-quantized representations to turn the task of future prediction into a next token prediction task, a technique that has been successfully employed in large language models. GAIA-1 has demonstrated its capability to acquire a comprehensive understanding of the environment, distinguishing between various concepts such as cars, trucks, buses, pedestrians, cyclists, road layouts, buildings, and traffic lights --- all through self-supervision. Further, GAIA-1 harnesses the capabilities of video diffusion models to generate realistic driving scenarios, thereby functioning as an advanced neural simulator. GAIA-1 is a multimodal approach that enables the control of the ego-vehicle's actions and other scene attributes through a combination of textual and action-based instructions.

While our method demonstrated promising results that have the potential to push the boundaries of autonomous driving, it is important to acknowledge current limitations. For instance, the autoregressive generation process, while highly effective, does not yet run at real-time. Nevertheless, it is noteworthy that this process lends itself well to parallelization, allowing for the concurrent generation of multiple samples.

The significance of GAIA-1 extends beyond its generative capabilities. World models represent a crucial step towards achieving autonomous systems that can understand, predict, and adapt to the complexities of the real world. Furthermore, by incorporating world models into driving models, we can enable them to better understand their own decisions and ultimately generalize to more real-world situations. Lastly, GAIA-1 can also serve as a valuable neural simulator, allowing the generation of unlimited data, including adversarial examples, for training and validating autonomous driving systems. 

\newline
\newline
\newline
\newline
\begin{ack}
\begin{small}
This work was made possible through the expertise and generous help of many teams and people across Wayve. In particular we would like to thank: Giulio D'Ippolito, Dan Reisman, Alex Persin, Przemyslaw Mazur, Oleg Sinavski, Long Chen, Fergal Cotter, Corina Gurau, Shu Ishida, Remi Tachet, Rudi Rankin, Tilly Pielichaty, Rod Bauer, Charlie Lyons-Rothbart, Harriett-Rose Follas, Robert Weston, Becky Goldman, Sasha Harrison, Saurabh Nair, Prajwal Chidananda, Tom Newton, Benoit Hanotte, Ana-Maria Marcu, Thomas Sajot, Giacomo Gallino, Alex Garcia Mayans, Tim Geypens, Robin Tweedie, Rebecca Hills, Tim Williams-Silvera, Darren Jenner, Matt Wood, Dave Chilvers, Danny Ly, Joseph Rodrigo, Will Dias, Naomi Standard, and Theepa Balasubramaniam.
\end{small}
\end{ack}

\clearpage
\bibliography{bibliography}

\end{document}